\documentclass[letterpaper]{article} % DO NOT CHANGE 
\usepackage{aaai2026}  % DO NOT CHANGE THIS
\usepackage{times}  % DO NOT CHANGE THIS
\usepackage{helvet}  % DO NOT CHANGE THIS
\usepackage{courier}  % DO NOT CHANGE THIS
\usepackage[hyphens]{url}  % DO NOT CHANGE THIS
\usepackage{graphicx} % DO NOT CHANGE THIS
\usepackage{natbib}  % DO NOT CHANGE THIS AND DO NOT ADD ANY OPTIONS TO IT
\usepackage{caption} % DO NOT CHANGE THIS AND DO NOT ADD ANY OPTIONS TO IT
\usepackage{algorithm}
\usepackage{algorithmic}

\usepackage{newfloat}
\usepackage{listings}
\DeclareCaptionStyle{ruled}{labelfont=normalfont,labelsep=colon,strut=off} % DO NOT CHANGE THIS
\floatstyle{ruled}
\newfloat{listing}{tb}{lst}{}
\floatname{listing}{Listing}
\def\eqref#1{Eq.~\ref{#1}}
\def\1{\bm{1}}

\usepackage{multirow,tabularx}
\usepackage{adjustbox}
\usepackage{enumitem}
\usepackage{amsmath}
\usepackage{amsthm}

\usepackage{subcaption}
\newtheorem{theorem}{Theorem}

\newtheorem{proposition}[theorem]{Proposition}
\usepackage[utf8]{inputenc} % optional, often already included
\usepackage{booktabs}
\usepackage{amsfonts}
\usepackage{nicefrac}
\usepackage{microtype}
\usepackage{xcolor} % okay unless used in text (avoid if unsure)
\usepackage{amssymb}
\usepackage{pifont}
\newcommand{\circledOne}{\ding{172}}
\newcommand{\circledTwo}{\ding{173}}

\title{Structural Disentanglement of Causal and Correlated Concepts}

\author{
  Qilong Zhao\thanks{Both authors contributed equally to this work.}, Shiyu Wang\footnotemark[1], Zeeshan Memon, Yang Qiao, \\
  Guangji Bai, Bo Pan, Zhaohui Qin, Liang Zhao\thanks{Corresponding author.} \\
}
\affiliations{
    Emory University\\
    Atlanta, GA, USA\\
    liang.zhao@emory.edu
}

\nocopyright
\begin{document}

\maketitle
\maketitle
\maketitle
\maketitle
\maketitle
\maketitle
\maketitle
\maketitle

\begin{abstract}
Controllable data generation aims to synthesize data by specifying values for target concepts. Achieving this reliably requires modeling the underlying generative factors and their relationships. In real-world scenarios, these factors exhibit both causal and correlational dependencies, yet most existing methods model only part of this structure.
We propose the \textbf{\underline{C}}ausal-\textbf{\underline{C}}orrelation \textbf{\underline{V}}ariational \textbf{\underline{A}}uto\textbf{\underline{e}}ncoder (\textbf{C$^2$VAE}), a unified framework that jointly captures causal and correlational relationships among latent factors. C$^2$VAE organizes the latent space into a structured graph, identifying a set of root causes that govern the generative processes. By optimizing only the root factors relevant to target concepts, the model enables efficient and faithful control.
Experiments on synthetic and real-world datasets demonstrate that C$^2$VAE improves generation quality, disentanglement, and intervention fidelity over existing baselines.
\end{abstract}


\begin{figure*}[!t]
\begin{center}
\includegraphics[width=1\textwidth]{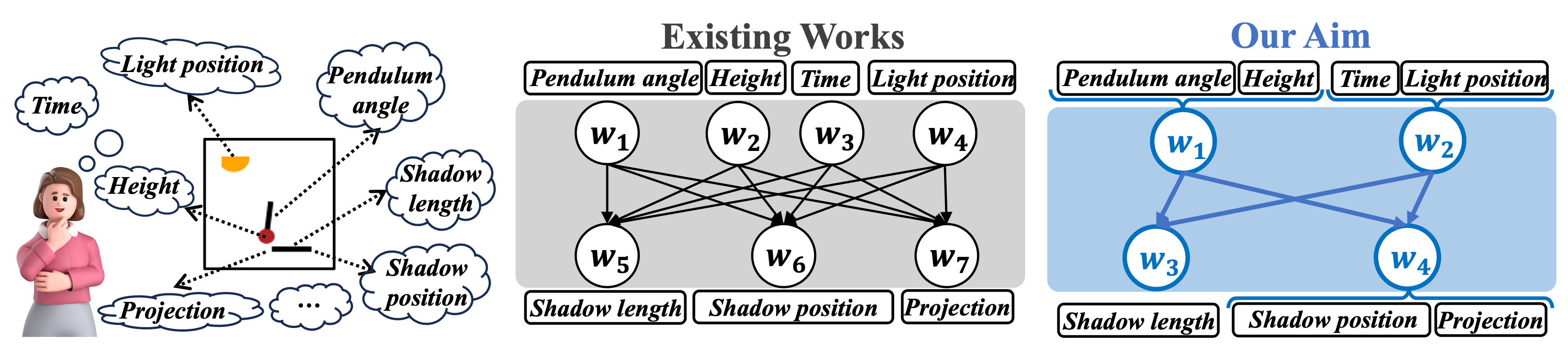}
\caption{A light source illuminates a swinging pendulum, casting a shadow on the floor. Observers can report many \emph{concepts} (e.g., \textit{Pendulum angle}, \textit{Light position}, \textit{Height}, \textit{Projection}, \textit{Shadow position}, \textit{Shadow length}, and \textit{Time}), but these often overlap and only partially reflect the true generative factors. For example, neither \textit{Height} nor \textit{Pendulum angle} alone provides a complete description of the pendulum's motion. Existing causal generative models often struggle with such spurious or redundant relationships among observed concepts. We aim to uncover a compact set of underlying \emph{latent factors} and the causal structure among them. In this example, $w_1$ and $w_2$ (shown in blue) are \emph{root factors}—independent sources that drive all other latent factors and, through them, the observed concepts.}
\label{fig:example}
\end{center}
\end{figure*}

\section{Introduction}
\label{sec:intro}

Generative models learn compact, low‑dimensional representations of complex, high‑dimensional observations. Such representations expose the generative mechanisms that produce the data and have driven rapid progress across natural language processing~\cite{zhang2023survey, jin2024amazon}, molecular design~\cite{xu2023geometric, born2023regression}, and image synthesis~\cite{epstein2024diffusion, wang2022multi}. Beyond improving likelihoods, low‑dimensional latents improve enable smooth interpolation and editing, and provide a practical basis for controllable generation.

In controllable generation, we synthesize data by setting values for human‑interpretable properties (which we will call concepts) via the model’s latent representation~\cite{klys2018learning, kingma2014semi, guo2020property}. To make such control reliable, the latent representation should (i) disentangle the underlying generative factors that produce the data and (ii) encode how these factors relate to one another. In real-world systems, factors interact in two distinct ways: some stand in causal relations (deliberately changing one factor leads to a predictable change in another), while others exhibit correlation (co‑variation due to shared influences) without implying causation. Ignoring this structure yields entangled controls and spurious generalization.
Given that both causal and correlational relationships exist among factors, existing causal generative models~\cite{yang2021causalvae, shen2022weakly} address only part of the structure by focusing on causal links while largely overlooking non-causal dependencies induced by (unobserved) shared influences.

Consider a system illustrated in Figure~\ref{fig:example}: a light source, a swinging pendulum, and its shadow. Sensors/observers may report many concepts, such as \textit{Pendulum angle}, \textit{Light position}, \textit{Height}, \textit{Projection}, \textit{Shadow position}, \textit{Shadow length}, and \textit{Time}. Several of these are redundant measurements of the same underlying factors (e.g., geometry and motion), and building a graph directly over observed concepts can misrepresent the true mechanism. A faithful approach should instead recover a structured set of latent factors along with their directed effects and any non‑causal dependencies among them.

In this work, we address the fundamental question:
\textit{``How can generative models capture the underlying mechanisms of data generation?''}
Our goal is controllable generation that respects the actual generative process—capturing both the causal structure among latent factors (the factors that produce the data) and the correlational dependencies that arise from shared influences. This is challenging for two main reasons. First, while causal relationships often produce correlations, correlation does not imply causation~\cite{pearl2009causality}. Second, real-world data often contain complex, overlapping effects, making it difficult to distinguish direct causal influence from mere co-variation.

We propose \textbf{\underline{C}}ausal-\textbf{\underline{C}}orrelation \textbf{\underline{V}}ariational \textbf{\underline{A}}uto\textbf{\underline{e}}ncoder (\textbf{C$^2$VAE}), a generative framework that models both causal and non-causal (correlational) relationships among latent factors. The model factorizes the latent space into root factors—factors with no parents in the latent causal graph, and their descendants, enabling effective control. Once trained, C$^2$VAE supports targeted concept manipulation via a joint objective (optimizing multiple concept targets simultaneously): rather than directly constraining every desired concept, we identify the relevant root factors that drive them and optimize only those. This strategy reduces control complexity while keeping generated samples consistent with the learned causal and correlational structure of the data.

% \vspace{-4mm}

\section{Problem Formulation}
\label{sec:problem}

Given a dataset $\mathcal{D} = \{(x^{(i)}, y^{(i)})\}_{i=1}^N$, where $x \in \mathcal{X}$ denotes high-dimensional observations (e.g., images), and $y \in \mathbb{R}^m$ is a vector of observed, human-interpretable \textit{targeted concepts} (e.g., pendulum angle, shadow length). These concepts represent quantities we wish to control. Importantly, they are derived from, but do not correspond one-to-one with the underlying generative factors that drive the system (i.e., the process that produces the observed data). Instead, they are often noisy, redundant, and entangled.

We assume $(x,y)$ are generated by some random processes from continuous latent variables $(w,z)$, where: $w = \{w_1, ..., w_n\}$ are the underlying variables responsible for generating both $x$ and $y$. These follow a structural causal model (SCM) with a directed acyclic graph (DAG) $\mathcal{A}$, encoding their causal relationships;
$z = \{z_1, ..., z_k\}$ are latent variables that influence $x$ but are assumed independent of both $w$ and $y$, i.e., $z \perp\!\!\!\perp (w, y)$. We do not supervise $z$ with concept labels during training, so $z$ is unlikely to encode information predictive of the targeted concepts; instead, it absorbs residual variation in $x$ (e.g., background, texture) unrelated to concept-level semantics~\cite{guo2020property}.

Within $w$, two kinds of relations exist.
\textbf{(1) Direct causal effects:}
      For latent variables $w_i,w_j$, we say $w_i$ is a \emph{direct cause}
      of $w_j$ iff the SCM~\cite{pearl2009causality} contains an arrow $w_i\!\to\!w_j$—equivalently,
      there exist values $a\neq a'$ such that
      $P(w_j\mid\mathrm{do}(w_i{=}a)) \neq P(w_j\mid\mathrm{do}(w_i{=}a'))$.
\textbf{(2) Correlation (non-causal dependence):}
      Two variables are \emph{correlated} when they are statistically
      dependent,
      $P(w_i,w_j) \neq P(w_i)\,P(w_j)$.
      This may arise from a direct causal link, a shared ancestor, or other open paths. 

The generative process is: $\quad x \sim p_\theta(x \mid w, z), \quad y \sim p_\gamma(y \mid w),$
where both $p_\theta(x \mid w, z)$ and $p_\gamma(y \mid w)$ are parameterized as nonlinear, stochastic mappings via neural networks.
We define \emph{root factors} as $w_{\text{root}} = \{w_i \mid \operatorname{pa}(w_i) = \varnothing\}$, where $\mathrm{pa}(w_i)$ denotes the parents of $w_i$ in $\mathcal{A}$. Manipulating only $w_{\text{root}}$ suffices to determine all causal variables $w$, and hence, the targeted concepts $y$ through $p_\gamma(y \mid w)$. This enables faithful and efficient concept control.

Our objective is therefore to learn a latent representation that captures \emph{both} the directed (causal) edges and the undirected (correlational) dependencies among the variables $w$ in a single, unified framework.

\section{Related Work}
\label{sec:bg}
We briefly summarize key approaches in controllable generation and causal representation learning; a comprehensive review is in Appendix.

Controllable generation has evolved from methods focusing on independent concept control (SemiVAE~\citep{locatello2019disentangling}, CSVAE~\citep{klys2018learning}, PCVAE~\citep{guo2020property}) to those addressing either causal relationships (CausalVAE~\citep{yang2021causalvae}, DEAR~\citep{shen2022weakly}) or correlations (CorrVAE~\citep{wang2022multi}). In parallel, causal representation learning has developed various techniques, including SCM-based methods~\citep{shimizu2006linear, yang2021causalvae}, domain generalization approaches~\citep{kong2023partial}, and latent variable causal discovery~\citep{cai2019triad}. Despite these advances, no existing approach simultaneously addresses both causal and correlational relationships in a unified framework, which is the primary contribution of our work.

\section{Causal-Correlation Variational Autoencoder}
\label{sec:method}

\begin{figure*}
\begin{center}
\includegraphics[width=0.9\textwidth]{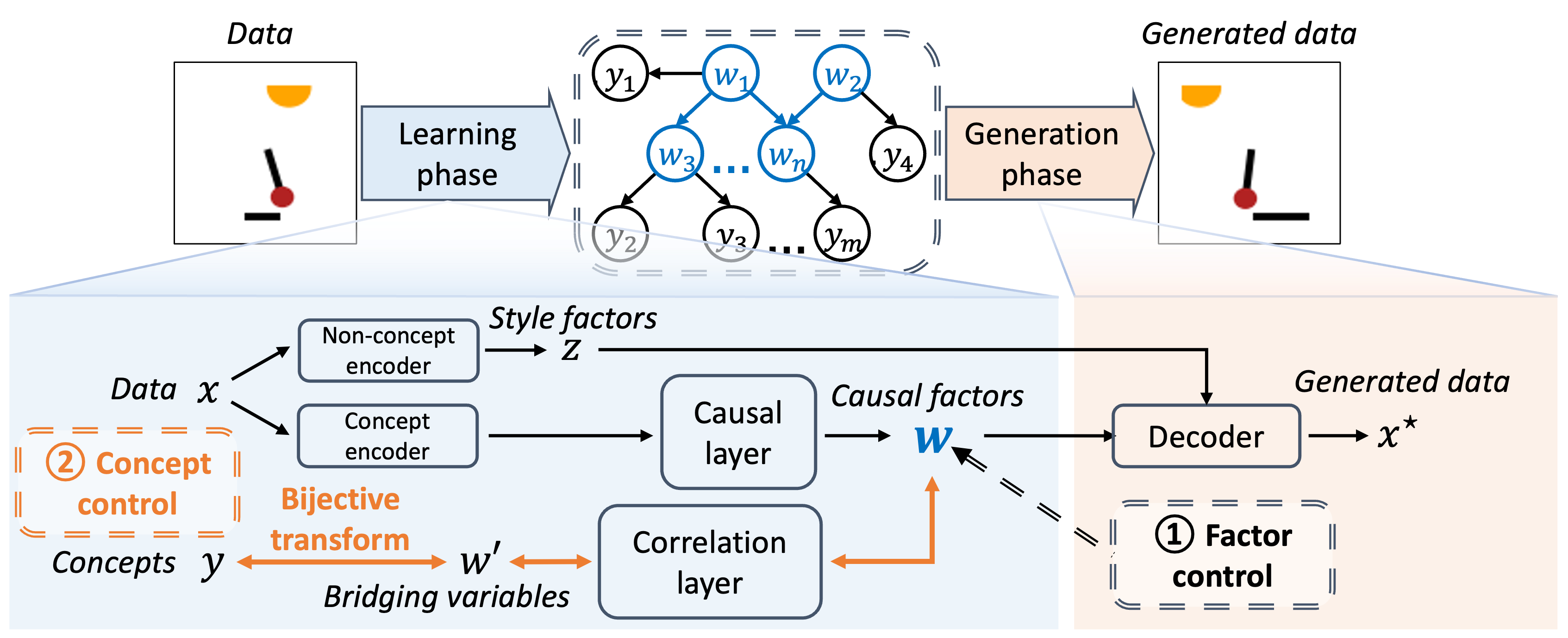}
\caption{
Overview of C$^2$VAE, which consists of a \emph{learning phase} and a \emph{generation phase}.  
During the learning phase, a \textit{Causal Layer} infers the structural causal graph among factors~\(w\), while a \textit{Correlation Layer} captures correlation between concepts~\(y\) and ~\(w\).  
In the generation phase, we provide two complementary modes of control:
\textbf{\circledOne\ Factor control}: edit the root factors \(w_{\text{root}}\) (a subset of factors \(w\)) to steer downstream factors, and thus the desired \(x^{\star}\).
\textbf{\circledTwo\ Concept control}: specify target concepts \(y^{\star}\); a bijective mapping propagates these targets back to the corresponding root factors \(w_{\text{root}}\), enabling the decoder to generate an image \(x^{\star}\) with \(y^{\star}\).
}
\label{fig:overview}
\end{center}
\end{figure*}

To address the problem formulated above and tackle the challenges, we propose the C$^2$VAE framework shown in Figure~\ref{fig:overview}. Our approach operates in two distinct phases: \textbf{(1) Learning phase}, where we encode data into a structured latent space that captures both causal and correlational relationships among concepts through a structural causal model and correlation layer; and \textbf{(2) Generation phase}, where we leverage the identified root factors ($w_{\text{root}}$) to enable controlled generation of new data with desired concept values via either direct factor manipulation or target concept specification.

\textbf{During learning phase}, the image \( x \) is first encoded into causal endogenous factors \( w \) that correspond to causally related concepts in data through a learned causal graph. Next, we incorporate a mask pooling layer to reveal the correlations among concepts \( y \). As a unified framework, C$^2$VAE optimizes only the ``root'' latent factors (\( w_{\text{root}} \)), which encode both causal and correlational relationships among concepts, ensuring efficient and interpretable control over concept generation.

\textbf{During the generation phase}, C$^2$VAE enables two complementary modes of control:
\textbf{\circledOne\ Factor control:} By directly editing the root factors \(w_{\text{root}}\) (a subset of factors \( w \)) , we can systematically steer all downstream factors to generate the desired output \(x^{\star}\), ensuring the generated data exhibits the desired concepts \(y^{\star}\) while maintaining consistency with the underlying causal and correlational structure.
\textbf{\circledTwo\ concept control:} Users can directly specify target concepts \(y^{\star}\), which our model maps back to the corresponding root factors \(w_{\text{root}}\) through a bijective transformation. The decoder then produces an image \(x^{\star}\) that manifests these target concepts. For simultaneous manipulation of multiple concepts, C$^2$VAE employs multi-objective optimization that focuses exclusively on the root factors \(w_{\text{root}}\) to ensure computational efficiency and causal consistency.

\subsection{Learning phase}

\subsubsection{Overall learning objective based on probabilistic generative model.}
Based on the problem formulation, the goal of C$^2$VAE is not only to model the generation of $x$ from $(w, z)$ but also to model the dependence between $y$ and $w$ for controlling targeted concepts. Therefore, we propose to achieve this goal by maximizing the joint log-likelihood $p(x, y)$ via its evidence lower bound (ELBO):
\begin{align}
    \log p(x, y) 
    & = \log \mathbb{E}_{q(w, z|x, y)}\left[\frac{p(x, y, w, z)}{q(w, z|x, y)}\right] \nonumber \\
    &
    \geq \, \mathbb{E}_{q(w, z|x, y)}\left[\log \frac{p(x, y, w, z)}{q(w, z|x, y)}\right] \nonumber \\
    & = \, \mathbb{E}_{q(w, z|x,y)}\big[\log p(x|w, z) + \log p(y|w) \nonumber \\
    & + \log p(w, z) - \log q(w, z|x, y)\big]
\end{align}

The last equation in the above is based on the decomposition of $\log p(x, y, w, z)$ into $\log p(x, y|w, z)+\log p(w, z)$, and the following two assumptions: (1) $x$ and $y$ are conditionally independent given $w$, as $w$ only captures information from $y$; (2) $z$ are assumed independent of $(w,y)$; hence $y\perp\!\!\!\perp z \mid w$. Hence, the overall learning objective of C$^2$VAE can be written as:
\begin{align}
    \mathcal{L}_1 = & - \mathbb{E}_{q(w, z|x, y)}\left[\log \frac{p(x, y, w, z)}{q(w, z|x, y)}\right] \nonumber \\
    = & - \mathbb{E}_{q(w, z|x)}\left[\log p_\theta(x|w, z)\right] 
    - \mathbb{E}_{q(w|x)}\left[\log p_\gamma(y|w)\right] \nonumber \\
    & + D_{KL}\big(q_\phi(w,z|x, y) \| p(w, z)\big),
    \label{eq:l1}
\end{align}
where $\theta$, $\gamma$, and $\phi$ are the parameters of the decoder, concept predictor (i.e., an MLP), and encoder, respectively. The final term in the objective function of Eq.~\ref{eq:l1} neither enforces the independence of $z$ from both $w$ and $y$ nor ensures that the components of $w$ are independent of each other. To address this, we further decompose it into the following KL terms:
\begin{align}
\label{eq:kl}
    \mathcal{L}_2 =& \rho_1\cdot D_{KL}(q_{\phi_1}(w|x, y)q_{\phi_2}(z|x)\|p(w, z)) \nonumber \\
    +&\rho_2\cdot D_{KL}(q_{\phi}(w, z|x, y)\|q_{\phi_1}(w|x)q_{\phi_2}(z|x)),
\end{align}
where $\rho_1$ and $\rho_2$ are hyperparameters to penalize two terms. $\phi_1$ and $\phi_2$ are model parameters of the concept encoder and the non-concept encoder, respectively. A sensitivity analysis of $\rho_1$ and $\rho_2$ is provided. 

\subsubsection{Causal Disentanglement Learning.}
To model directed dependencies among $w=\{w_1,\dots,w_n\}$, we use a linear
structural causal model (SCM) as in~\cite{yang2021causalvae}:
\begin{align}
    w \;=\; A^\top w + \varepsilon \;=\; (I - A^\top)^{-1}\varepsilon,
    \qquad \varepsilon \sim \mathcal{N}(0, I),
    \label{eq:A}
\end{align}
where the components of $\varepsilon$ are independent and $A$ is the adjacency
of a DAG (acyclicity enforced by a differentiable constraint; equivalently,
some permutation makes $A$ strictly upper triangular). We parameterize $A$ and
learn it jointly with the rest of the model.

\subsubsection{Correlation-aware Conditional Prior.}
\label{sec:correlation prior}
\textbf{Let} $m$ \textbf{denote the number of} \emph{observed concepts} (i.e., $y \in \mathbb{R}^m$) \textbf{and} $n$ \textbf{the number of} \emph{causal latent variables} (i.e., $w \in \mathbb{R}^n$). \\
As we supposed, the correlation between concepts is unknown, so $\Omega_i(y)$ (i.e., $i=1, 2, ..., n$) has to be learned by the model. To achieve this goal, we use a mask pooling layer $M\in\{0, 1\}^{n\times m}$. $M$ captures how $w$ correlates with $y$, where $M_{ij}=1$ means $w_i$ correlates with the $j$-th concept $y_j$; otherwise, there is no correlation. In the meantime, to ensure the identifiability of $M$, we mask all its upper triangular values, enforcing a hierarchical structure that prevents symmetric dependencies and uniquely defines concept correlations. Thereafter, two variables that connect to the same variables in $w$ are regarded as correlated. The binary elements in $M$ are made trainable via the Gumbel Softmax function, while the $L_1$ norm of the mask matrix is added to the learning objective to encourage sparsity, reducing redundant relationships and improving interpretability.

Given $M$, we are able to collect the correlated concepts $\Omega_i(y)$ as follows:
\begin{align}
    \Omega_i(y)=\{y_j: M_{ij}=1, i\ge j\}
\end{align}
Specifically, for concepts $y$, we can compute the corresponding $w'$ that projects values from $w$ to each concept as $w\cdot J^T\odot M$, where each column corresponds to the related latent variables in $w$ to predict the corresponding concept in $y$. Here, $J$ is a vector with all values set to one, and $\odot$ represents element-wise multiplication. For each concept $y_j$ in $y$, we aggregate all the information from its related latent variable set in $w$ into the next-level bridging variable $w_j'$ via the aggregation function $a$:
\begin{align}
    w'=a(w\cdot J^T\odot M; \beta),
\end{align}
where $\beta$ is the parameter of the aggregation function $a$. Finally, $y$ can be predicted using $w'$ as:
\begin{align}
    y=f_c(w';\gamma),
    \label{eq:y_pred}
\end{align}
where $\gamma$ are the parameters of a set of concept predictors $f_c$. The proposed framework aims to generate data that preserves the specific desired concept values. Therefore, it is natural to model the dependence between $w_j'$ and $y_j$~\citep{wang2022multi}. To ensure bijectivity and enable reliable inversion, $f_c$ is constrained by a \textit{Lipschitz-constant}~\cite{behrmann2019invertible}. This guarantees stable transformations, allowing for precise, controllable generation with specified concept values:
\begin{align}
    \mathcal{L}_3 = & - \sum_{j=1}^m \mathbb{E}_{w' \sim p(w')}\left[\mathcal{N}(y_j | f_c(w_j'; \gamma_j))\right] \nonumber \\
    & + \left\| Lip(f_c(w_j'; \gamma_j)) - 1 \right\|_2
    \label{eq:l3}
\end{align}

\subsection{Generation phase}
\label{sec:inf}
In this section, we explain the process of generating data with specified concept values $y^{\star}$, which corresponds to the \textbf{\circledTwo\ concept control} mode illustrated in Figure~\ref{fig:overview}. This approach allows direct manipulation of targeted concepts.

The first step is to generate $w'$ given $y^{\star}$. As we have learned an invertible mapping between $w'$ and $y$ in Eq.~\ref{eq:l3}, it is necessary to back-compute $w'$ given $y^{\star}$ by solving:
\begin{align}
    w' & = \arg\max_{w'} \log p(y = y^{\star} | w') \nonumber \\
    &
    =  \arg\max_{w'} - \sum_{j=1}^m \left(y^{\star}_j - f(w_j'; \gamma_j)\right)^2
\end{align}
Next, given $w'$, we aim to infer $w$. We naturally formalize this process of searching for $w^{\star}$ that satisfies specific concept values as a multi-objective optimization framework. Note that we have:
\begin{align}
    p(w|w')\propto p(w, w')=p(w'|w) p(w)=p(y|w) p(w),
    \label{eq:bayes}
\end{align}
given that $y$ is the bijective mapping from $w'$ via an invertible function, the first term of Eq.~\ref{eq:bayes} is the likelihood to predict $y$. The second term is the prior distribution of $w$, which is $\mathcal{N}(0,I)$. Additionally, to reduce the complexity of the optimization process, we leverage the causal relationships between concepts to minimize the number of variables to be optimized in $w$. Specifically, we only use the \emph{root factors} in the adjacency matrix $A$ learned by Eq.~\ref{eq:A}, by masking other factors: $w_{root}=(\sum_{i=1}^n a_i)\odot w$, where $a_i$ is the vector of the $i$-th column of $A$. Therefore, the learning objective is formulated as:
\begin{align}
    w^{\star} \Longrightarrow & \arg\max_{w_{\text{root}} \sim p(w_{\text{root}})} \left(\log p(y | w_{\text{root}}) + \log p(w_{\text{root}})\right) \nonumber \\
    \text{s.t.,} \quad & y = y^{\star}
\end{align}
The above optimization is achieved in a weighted-sum manner using general learning algorithms such as stochastic gradient descent or Adam~\citep{kingma2014adam}. The learned $w$ is combined with the sampled $z$ to serve as the input of the data decoder, which then generates the data.

\begin{table*}[hbt!]
 \caption{Image generation quality measured by Frechet Inception Distance (FID ↓) and Peak-Signal-to-Noise Ratio (PSNR ↑).}
% \vspace{-3mm}
\centering
\begin{tabular}{lcccc ccc ccc ccc}
        \hline
        \multirow{2}{*}{\textbf{Method}} && \multicolumn{2}{c}{\textbf{Pendulum}} && \multicolumn{2}{c}{\textbf{Flow}} && \multicolumn{2}{c}{\textbf{dSprites}} && \multicolumn{2}{c}{\textbf{CelebA}} \\
        \cmidrule(r){3-4} \cmidrule(lr){6-7} \cmidrule(lr){9-10} \cmidrule(l){12-13}
        && \textbf{FID ↓} & \textbf{PSNR ↑} && \textbf{FID ↓} & \textbf{PSNR ↑} && \textbf{FID ↓} & \textbf{PSNR ↑} && \textbf{FID ↓} & \textbf{PSNR ↑} \\
        \hline
        SemiVAE && 33.83 & 14.96 && 28.79 & 23.96 && 74.47 & 12.08 && 238.36 & 8.05 \\
        CSVAE && 40.26 & 14.97 && 29.05 & 23.97 && \textbf{51.01} & 12.13 && 245.31 & 8.53 \\
        PCVAE && 28.17 & 14.93 && 29.69 & 24.07 && 95.94 & 12.19 && 235.98 & 8.11 \\
        CorrVAE && 32.23 & 14.97 && 29.83 & 24.01 && 104.11 & 12.23 && 235.95 & 7.98 \\
        ICM-VAE && 78.99 & \textbf{27.15} && 28.92 & 24.38 && 61.62 & \textbf{25.59} && 149.27 & 15.28 \\
        CausalVAE && 31.04 & 19.38 && 11.10 & 26.12 && 166.97 & 22.56 && 161.61 & 17.60 \\
        \hline
        \textbf{C$^2$VAE-true} && \textbf{10.60} & 23.33 && \textbf{5.51} & \textbf{28.78} && 109.82 & 19.73 && \textbf{118.63} & \textbf{18.51} \\
        \textbf{C$^2$VAE} && 25.81 & 20.14 && 8.81 & 26.56 && 95.24 & 21.53 && 165.04 & 17.15 \\
        \hline
    \end{tabular}
\label{tab:eval}
\end{table*}

\section{Identifiability Analysis}
\label{sec:identifiability}

We address two questions: (i) identifiability of the concept-bearing latents $w$ from $(x,y)$, and (ii) identifiability of structure (the latent DAG $A$ up to observational limits and the correlation mask $M$). We adopt the standard iVAE assumptions and notation \cite{khemakhem2020variational}; in particular, $u\equiv y$ serves as the auxiliary (i.e., concepts $y$ serve as the auxiliary variable), the conditional prior $p(w\!\mid\!y)$ is factorial exponential-family with the iVAE rank/variability condition, and the decoder $(w,z)\mapsto x$ is injective (e.g., invertible/flow or injective with small output noise).

\subsection{Concept-latent identifiability}
\begin{theorem}[identifiability of $w$]
\label{thm:ivae}
Under the iVAE assumptions with $u\equiv y$, the latent vector $w$ is identifiable from $p(x,y)$ up to $\sim_{\textsc{iVAE}}$. In particular, when each $p(w_i\!\mid\!y)$ is modulated by sufficiently rich sufficient statistics (e.g., mean and variance) and the variability condition holds, the residual linear ambiguity collapses to a permutation. No identifiability is claimed for the nuisance $z$ (which follows an unconditional prior and satisfies $z\!\perp\!(w,y)$ in the generative model).
\end{theorem}

\subsection{Causal and correlational structure}
\paragraph{Causal DAG (observational limits).}
As in the technique section, $w=(I-A^\top)^{-1}\varepsilon$ is the linear SCM with mutually independent disturbances. From $p(x,y)$, the observational linear-Gaussian case identifies only the Markov equivalence class of $A$. Stronger asymmetry (e.g., non-Gaussian independent disturbances, equal-error-variance, or multi-environment variance diversity) upgrades to full DAG identifiability; we do not impose such assumptions here.

\paragraph{Correlation mask (concept–latent support).}
To ensure hierarchical structure and identifiability, we use only mild design constraints specific to the correlation layer: $M\in\{0,1\}^{n\times m}$ is lower-triangular with ones on the diagonal and $\ell_1$-sparse; the aggregator $a(\cdot)$ is permutation-invariant and strictly increasing in each argument; and the concept map $f_c$ is bijective (component-wise or triangular diffeomorphism), so $y\leftrightarrow w'$.

\begin{proposition}[Mask identifiability]
\label{prop:mask}
Under the above constraints, the parent set for each concept $y_j$,
$S_j=\{\,i : M_{ij}=1\,\}$, is uniquely determined from $p(x,y)$ up to the global permutation ambiguity allowed by $\sim_{\textsc{iVAE}}$; hence $M$ is identifiable.
\end{proposition}

\paragraph{Implications.}
Theorem~\ref{thm:ivae} provides disentanglement guarantees for $w$ (up to $\sim_{\textsc{iVAE}}$). Proposition~\ref{prop:mask} anchors concepts to specific subsets of $w$, making concept control predictable via the bijection $y\!\leftrightarrow\!w'$. For the causal layer, we remain within observational limits.

\begin{table}[t]
\centering
\caption{concept predictability (MAE ↓). MAE for dSprites and CelebA is shown as a percentage. MAE for ICM-VAE is not shown because it does not produce predicted concept values.}
\label{tab:mae}
\resizebox{\linewidth}{!}{
\begin{tabular}{lcccc}
        \hline
        \textbf{Method} & \textbf{Pendulum} & \textbf{Flow} & \textbf{dSprites} & \textbf{CelebA} \\
        \hline
        SemiVAE & 0.23 & 0.25 & 2.97 & 16.08 \\ 
        CSVAE & 8.99 & 12.16 & 7.57 & \textbf{0.22} \\ 
        PCVAE & 0.34 & 0.28 & 3.24 & 16.55 \\ 
        CorrVAE & 0.61 & 0.66 & 3.61 & 16.49 \\
        CausalVAE & 0.48 & 1.32 & 2.73 & 0.73 \\ 
        \hline
        \textbf{C$^2$VAE-true} & 0.65 & 0.44 & 2.66 & 17.30 \\
        \textbf{C$^2$VAE} & \textbf{0.18} & \textbf{0.19} & \textbf{2.43} & 15.27 \\ % Dummy values for CelebA
        \hline
\end{tabular}
}

\centering
\caption{Disentanglement measured by average Mutual Information (avgMI ↓) between learned representations and labels.}
\label{tab:avgmi}
\resizebox{\linewidth}{!}{
\begin{tabular}{lcccc}
        \hline
        \textbf{Method} & \textbf{Pendulum} & \textbf{Flow} & \textbf{dSprites} & \textbf{CelebA} \\
        \hline
        SemiVAE & 0.81 & 0.59 & 1.60 & 1.71 \\
        CSVAE & 1.55 & 1.00 & 1.04 & 0.85 \\
        PCVAE & 0.80 & 0.58 & 1.59 & 1.86 \\
        CorrVAE & 0.99 & 0.68 & 1.60 & 1.62 \\
        ICM-VAE & 0.84 & 1.49 & 1.52 & 2.24 \\ 
        CausalVAE & 1.05 & 0.97 & 0.98 & 1.28 \\ 
        \hline
        \textbf{C$^2$VAE-true} & 1.02 & 0.79 & \textbf{0.78} & 1.12 \\ 
        \textbf{C$^2$VAE} & \textbf{0.79} & \textbf{0.56} & 0.83 & \textbf{0.88} \\ 
        \hline
\end{tabular}
}
\end{table}

\section{Experiments}
\label{sec:exp}
In this section, we present the experimental results and discuss our findings. The proposed method is compared against existing state-of-the-art approaches for controllable data generation and causal generative modeling using four datasets (including both synthesized and real-world datasets). We assess the quality of the generated data, the capability of each method to learn causal and correlated representations, and whether the outcomes of interventions on the learned latent variables align with the expected correlations and causal relationships.

\begin{figure*}[hbt!]
\centering
\begin{subfigure}{.16\textwidth}
\includegraphics[width=\linewidth]{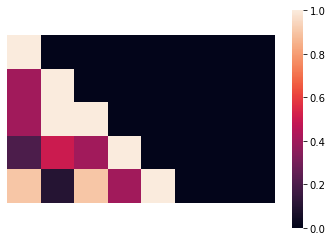}  
  \caption{epoch=1}
\end{subfigure}
\begin{subfigure}{.16\textwidth}
  \includegraphics[width=\linewidth]{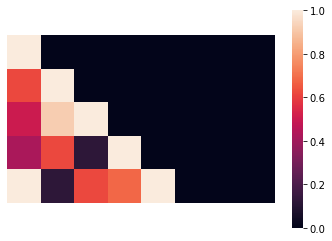}  
  \caption{epoch=5}
\end{subfigure}
\begin{subfigure}{.16\textwidth}
  \includegraphics[width=\linewidth]{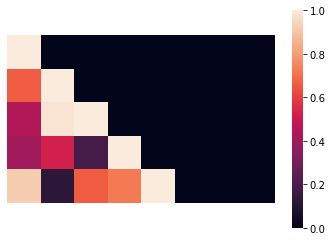}  
  \caption{epoch=10}
\end{subfigure}
\begin{subfigure}{.16\textwidth}
  \includegraphics[width=\linewidth]{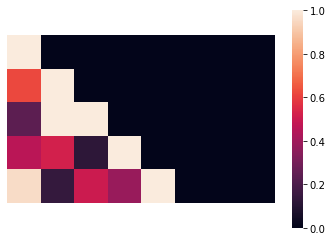}  
  \caption{epoch=50}
\end{subfigure}
\begin{subfigure}{.16\textwidth}
  \includegraphics[width=\linewidth]{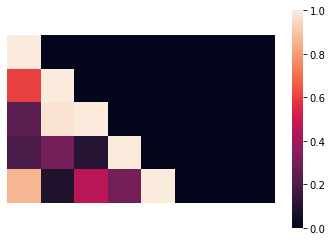}  
  \caption{epoch=100}
\end{subfigure}
\begin{subfigure}{.16\textwidth}
\includegraphics[width=\linewidth]{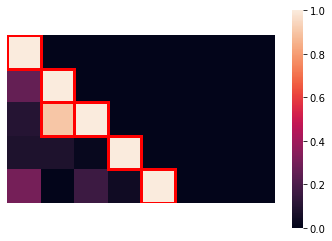}  
  \caption{epoch=200}
\end{subfigure}
\caption{The learning process of the mask pooling layer of C$^2$VAE. Rows correspond to \textit{Pendulum angle}, \textit{Light position}, \textit{Time}, \textit{Shadow length}, and \textit{Shadow position}, from the first to the last.}
\label{fig:mol_cond}
\end{figure*}

% \textbf{Figure 1: Visualizing mask pooling layer.}
\begin{figure*}[hbt!]
\centering
\begin{subfigure}{.16\textwidth}
\includegraphics[width=\linewidth]{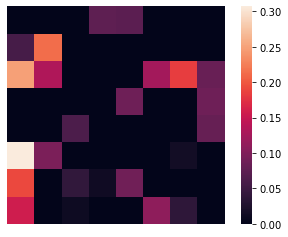}  
  \caption{epoch=1}
\end{subfigure}
\begin{subfigure}{.16\textwidth}
  \includegraphics[width=\linewidth]{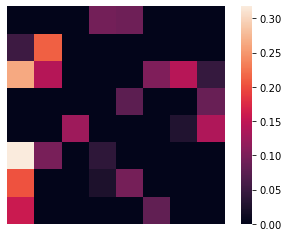}  
  \caption{epoch=5}
\end{subfigure}
\begin{subfigure}{.16\textwidth}
  \includegraphics[width=\linewidth]{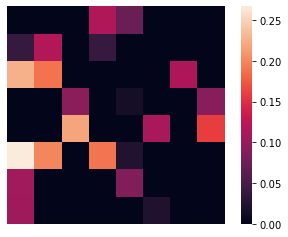}  
  \caption{epoch=10}
\end{subfigure}
\begin{subfigure}{.16\textwidth}
  \includegraphics[width=\linewidth]{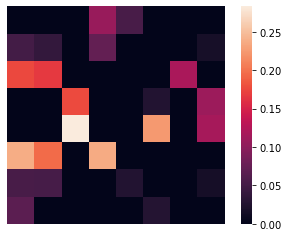}  
  \caption{epoch=50}
\end{subfigure}
\begin{subfigure}{.16\textwidth}
  \includegraphics[width=\linewidth]{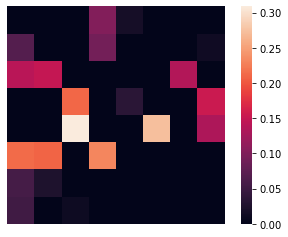}  
  \caption{epoch=100}
\end{subfigure}
\begin{subfigure}{.16\textwidth}
\includegraphics[width=\linewidth]{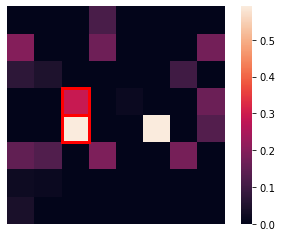}  
  \caption{epoch=200}
\end{subfigure}
\caption{The learning process of the causal graph of C$^2$VAE. In each subfigure, from left to right and top to bottom, the slots correspond to $w_1$ through $w_5$ and then $z_1$ through $z_3$. Here, $w_1$ to $w_5$ each represent a targeted concept: \textit{Pendulum angle}, \textit{Light position}, \textit{Time}, \textit{Shadow length} and \textit{Shadow position}; while $z_1$ to $z_3$ correspond to non-targeted factors.}
\label{fig:causal}
\end{figure*}

\subsection{Experimental Setup}

A more detailed introduction is in Appendix.

\textbf{Datasets.}
We evaluate on: (1) \textbf{Pendulum dataset}~\citep{yang2021causalvae}: 7k images with concepts (\textit{Pendulum angle}, \textit{Light position}) → (\textit{Shadow position}, \textit{Shadow length}), plus \textit{Time} correlated with \textit{Light position}. (2) \textbf{Flow dataset}~\citep{yang2021causalvae}: 8k images with concepts (\textit{Ball size} → \textit{Water height}, (\textit{Water height}, \textit{Hole}) → \textit{Water flow}). (3) \textbf{dSprites dataset}~\citep{dsprites17}: 730k images from which we derive 4 concepts (\textit{scale}, \textit{x position}, \textit{x+y position}, \textit{$x^2$+$y^2$ position}). (4) \textbf{CelebA dataset}~\cite{liu2015deep}: 200k face images with concepts \textit{Young}, \textit{Gender}, \textit{Receding hairline}, \textit{Makeup}, and \textit{Eye bag}.

\textbf{Baselines and Metrics.}
We compare against: SemiVAE~\citep{kingma2014semi}, CSVAE~\citep{klys2018learning}, PCVAE~\citep{guo2020property}, CorrVAE~\citep{wang2022multi}, ICM-VAE~\citep{komanduri2023learning}, and CausalVAE~\citep{yang2021causalvae}. We also include C$^2$VAE-true that uses ground-truth causal graphs and correlation masks as an ablation. We evaluate using: (1) FID~\citep{heusel2017gans} and PSNR for generation quality, (2) MAE for concept accuracy, and (3) avgMI~\citep{locatello2019disentangling} for disentanglement.

\subsection{Experimental Analysis}
We evaluate the proposed C$^2$VAE method through a series of experiments designed to answer the following research questions.
\textbf{RQ1:} How does C$^2$VAE perform in terms of overall data generation quality compared to existing methods? \textbf{RQ2:} Can C$^2$VAE accurately identify both correlation and causality between concepts? \textbf{RQ3:} How precisely can C$^2$VAE control causal and correlated concepts? \textbf{RQ4:} How effective are interventions on causal variables in C$^2$VAE?

To address these questions, we conducted comprehensive experiments across four datasets: Pendulum, Flow, dSprites, and CelebA. Below, we present our findings for each research question.

\begin{figure}[t]
  \centering
  \includegraphics[width=0.95\linewidth]{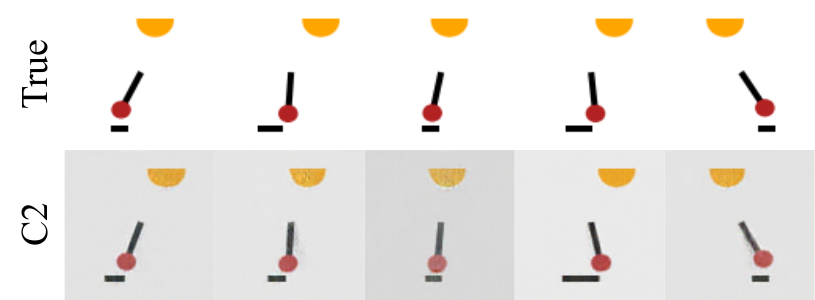}
\caption{\textbf{\circledTwo\ Concept control} on pendulum dataset by setting values for all concepts.}
\label{fig:cond_gen}
\end{figure}

\textbf{RQ1: Strong Overall Data Generation Quality.}
\textit{C$^2$VAE enhances performance on overall data generation quality despite its focus on controllable generation.}
As shown in Table~\ref{tab:eval}, C$^2$VAE achieves superior FID scores that are 14.94 (36.67\%), 17.42 (66.41\%), and 46.04 (21.81\%) better on average than other methods on Pendulum, Flow, and CelebA datasets, respectively. Additionally, its PSNR is 2.42 (13.66\%), 2.14 (8.77\%), and 6.22 (56.98\%) better on average across these datasets. When provided with ground truth causal graphs and correlation masks (C$^2$VAE-true), the model consistently achieves the strongest overall performance.
Interestingly, results on the dSprites dataset reveal a trade-off between concept control and data generation quality. Methods that excel in concept-centered evaluations often generate lower-quality data on this dataset. For example, CSVAE achieves the best FID score but the worst MAE, suggesting a compromise between causal performance and generation quality. Nevertheless, among methods with constraints on concept control or causal/correlation learning, C$^2$VAE still maintains strong data generation quality.

\textbf{RQ2: Identification Quality.}
\textit{C$^2$VAE successfully uncovers both correlation and causality between concepts in alignment with ground truth.}
Figure~\ref{fig:mol_cond} demonstrates how the correlation mask converges as training progresses. The model correctly identifies that $w_2$ controls both \textit{light position} and \textit{time}, indicating their correlation, which aligns with the ground truth relationship.
Similarly, Figure~\ref{fig:causal} shows that the causal graph learning also converges during training. The model correctly identifies that $w_3$ (\textit{Time}) causes $w_4$ and $w_5$ (\textit{Shadow length} and \textit{Shadow position}, respectively), which matches the true causal structure of these factors.

\textbf{RQ3: Precise concept Control.}
\textit{C$^2$VAE enables more precise control over both causal and correlated concepts than existing methods.}
Disentangled latent variables should each correspond to a concept of interest, forming a basis for discovering relationships and enabling specific interventions. Table~\ref{tab:mae} and Table~\ref{tab:avgmi} show that C$^2$VAE generally outperforms all other methods in terms of both MAE and avgMI, demonstrating superior performance in aligning latent variables with the concepts.
Specifically, the MAE of C$^2$VAE is 21.74\% better than the second-best method (SemiVAE) on Pendulum and 24.00\% lower on Flow. While the improvement in avgMI over SemiVAE is modest on Pendulum and Flow, it becomes substantial on dSprites, where the avgMI of C$^2$VAE is almost half that of SemiVAE, indicating much stronger concept disentanglement.
Figure~\ref{fig:cond_gen} further illustrates the concept control capabilities of C$^2$VAE. When generating images from the Pendulum dataset by specifying values for all concepts (using ground truth labels as $y^{\star}$), the corresponding ground truth image $x^{\star}$ is well recovered. This demonstrates that our model accurately captures the underlying generative mechanisms and enables faithful conditional generation based on the provided concept values.

\begin{figure}[t]
  \centering
  \includegraphics[width=0.95\linewidth]{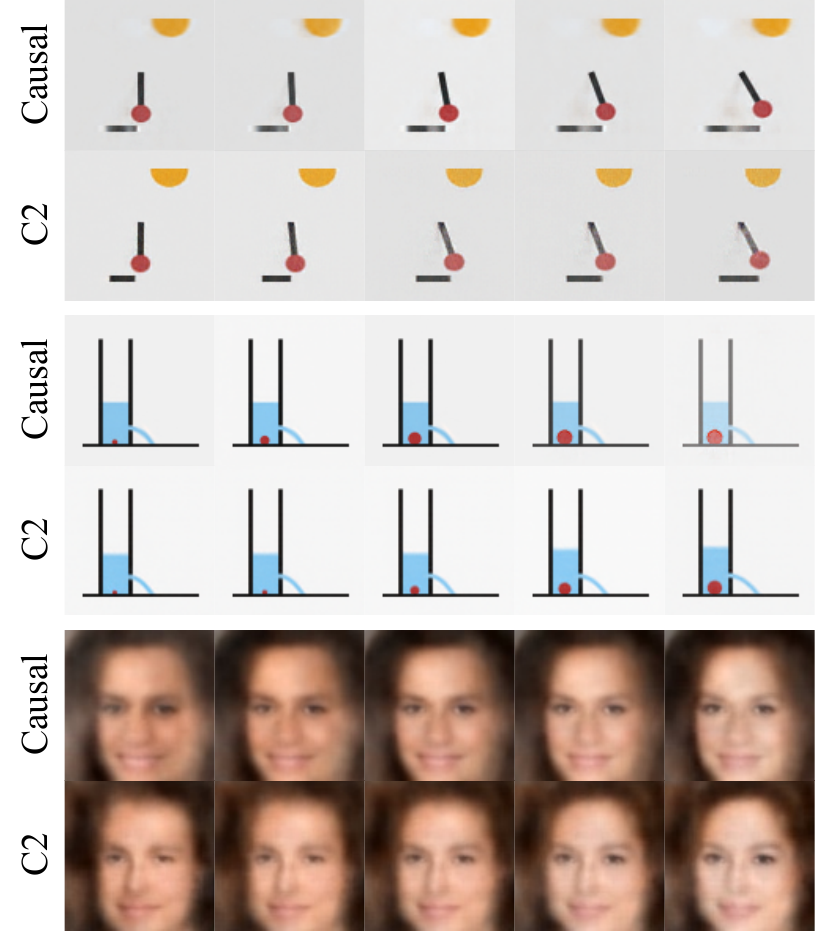}
\caption{\textbf{\circledOne\ Factor control} on \textit{Pendulum angle} in Pendulum dataset, \textit{Ball size} in Flow dataset, and \textit{Gender} in CelebA dataset.
}
\label{fig:causal_gen}
\end{figure}

\textbf{RQ4: Faithful Causal Interventions.}
\textit{C$^2$VAE achieves more realistic interventions that follow the true generative mechanisms compared to baselines.}
Figure~\ref{fig:causal_gen} demonstrates the effectiveness of intervening on single concepts across three datasets. The results reveal several advantages of C$^2$VAE over CausalVAE:
\textbf{Pendulum Dataset:} When modifying \textit{Pendulum angle}, CausalVAE generates images where the \textit{Shadow length} is often incorrectly longer than it should be (especially visible in the last column). In contrast, C$^2$VAE maintains accurate shadow properties that respect the physical relationship between pendulum angle and its shadow.
\textbf{Flow Dataset:} When modifying \textit{Ball size}, CausalVAE fails to properly adjust the \textit{Water flow} angle, which should change as a consequence of \textit{Water height} changing due to the ball size. C$^2$VAE correctly captures this causal chain, demonstrating its ability to model multi-step causal relationships.
\textbf{CelebA Dataset:} When intervening on \textit{Gender}, CausalVAE often inadvertently changes unrelated facial concepts like expressions. C$^2$VAE demonstrates more targeted modifications to gender-specific concepts (such as \textit{Hairstyle}) while preserving unrelated concepts like \textit{Young} and \textit{Eye bag}.

\textbf{Summary.} These findings confirm that C$^2$VAE achieves precise controllable generation over individual concepts while adhering to the true generative mechanisms, outperforming existing methods in respecting both causal and correlational structures during interventions.

\begin{table}[t]
\centering
\caption{Sensitivity analysis on KL weights $\rho_1$ and $\rho_2$ in Eq.~\eqref{eq:kl}. We report PSNR ↑, MAE ↓, and average mutual information (avgMI ↓) on the Pendulum dataset.}
\begin{adjustbox}{width=0.35\textwidth}
\footnotesize
\begin{tabular}{cccccc}
\toprule
$\rho_1$ & $\rho_2$ & PSNR $\uparrow$ & MAE $\downarrow$ & avgMI $\downarrow$ \\
\midrule
0        & 1        & 20.39           & 0.26             & 0.88 \\
1        & 0        & 20.73           & 0.42             & 1.63 \\
1        & 1        & 20.14           & \textbf{0.18}    & \textbf{0.79} \\
10       & 1        & 22.06           & 0.34             & 1.22 \\
1        & 10       & \textbf{22.36}  & 0.45             & 0.85 \\
0.1      & 1        & 21.19           & 0.39             & 0.81 \\
1        & 0.1      & 22.20           & 0.20             & 1.04 \\
\bottomrule
\end{tabular}
\label{tab:rho_ablation}
\end{adjustbox}
\end{table}

\subsection{Sensitivity Analysis}
\label{app:kl}
In Table~\ref{tab:rho_ablation}, the results reveal a trade-off between these objectives. \textbf{Setting $\rho_1=1$ and $\rho_2=1$ achieves the best balance}, producing optimal disentanglement (avgMI=0.79) and concept prediction accuracy (MAE=0.18), though at the cost of slightly lower reconstruction quality. Increasing $\rho_2$ to 10 while keeping $\rho_1=1$ yields the highest reconstruction quality (PSNR=22.36), but significantly degrades concept prediction accuracy (MAE=0.45). Conversely, setting either weight to 0 negatively impacts the model's ability to learn properly disentangled representations, as evidenced by higher avgMI values. Based on these findings, \textbf{we selected $\rho_1=\rho_2=1$ as our default configuration for all experiments}, prioritizing concept control capabilities while maintaining reasonable reconstruction quality—a choice that aligns with our primary goal of accurate causal and correlational concept learning.

\section{Conclusion}
\label{sec:conclusion}

We introduced \textbf{C}$^2$\textbf{VAE}, a unified variational framework that disentangles and jointly models \emph{causal} and \emph{correlational} concept structure for controllable data generation.  
By (i) learning a linear structural causal model over latent factors, (ii) discovering a sparse, lower-triangular correlation mask, and (iii) conditioning generation on a compact set of root causes, our method enables faithful interventions and multi-objective concept control while preserving the underlying mechanism.

\bibliography{aaai2026}
\appendix

\section{Comprehensive Related Work}
\label{app:related}

\subsection{Controllable Generation by Deep Learning}
Controllable generation by deep learning aims to synthesize data using deep generative models while preserving specific concepts~\citep{wang2024controllable}. Existing approaches can be categorized based on the complexity of relationships among the concepts to be controlled: 

\textbf{Independent concept control} focuses on learning dependencies between concepts and latent variables while ensuring disentanglement to maintain independence. Notable approaches include SemiVAE~\citep{locatello2019disentangling}, which factorizes the latent space to associate each latent variable with a specific concept; CSVAE~\citep{klys2018learning}, which captures binary concepts through independent latent variables with adversarial learning; and PCVAE~\citep{guo2020property}, which establishes invertible mutual dependence between concepts and latent variables for direct concept manipulation.

\textbf{Causal or correlated concept control} addresses the more challenging scenario of modeling inherent dependencies among concepts. CorrVAE~\citep{wang2022multi} extends PCVAE by incorporating a mask pooling layer to model concept correlations, framing data generation as a multi-objective optimization problem. For explicit causal modeling, CausalVAE~\citep{yang2021causalvae} enforces direct correspondence between latent variables and causal concepts through structural causal models (SCMs), while DEAR~\citep{shen2022weakly} relaxes concept supervision by introducing learnable mappings from concepts to latent variables.

Despite these advancements, jointly capturing both concept correlation and causality remains an open challenge, particularly given the complex dependencies in real-world data. This gap motivates our unified approach to causal and correlational concept control within a single generative framework.

\subsection{Causal Representation Learning}
Causal representation learning aims to uncover and represent underlying causal structures within data, with approaches broadly falling into several categories:

\textbf{SCM-based methods} learn causal graphs using structural causal models~\citep{shimizu2006linear, vowels2022d}, as applied in CausalVAE~\citep{yang2021causalvae} and DEAR~\citep{shen2022weakly}. Alternative approaches include causal regularization~\citep{bahadori2017causal}, which weights the learning objective by edge likelihoods, and attention-based methods that capture pairwise associations corresponding to linear-Gaussian SCMs~\citep{rohekar2024causal}. ICM-VAE~\citep{komanduri2023learning} models independent causal mechanisms using nonlinear flow-based functions supervised by causally related labels. 

\textbf{Domain generalization approaches} apply causal representation learning to improve cross-domain robustness by identifying domain-invariant causal factors~\citep{kong2023partial, lu2021invariant}, ensuring stability across diverse environments.

\textbf{Latent variable causal discovery} addresses scenarios where not all system factors are observed. Methods range from constraint-based approaches like Triad Constraints~\citep{cai2019triad}, which leverage independence conditions in pseudo-residuals, to rank-based techniques for identifying hierarchical structures~\citep{huang2022latent} and score-based extensions for latent-variable models~\citep{ng2024score}. 

\textbf{concept-based models (CBMs)} offer interpretable prediction through human-understandable units. The field has evolved from early approaches assuming concept independence, like concept Bottleneck Model~\citep{koh2020concept}, to integrating causal relationships (Causal concept Graph Models~\citep{dominici2024causal}), improving representation alignment (GlanceNets~\citep{marconato2022glancenets}), and providing identifiability guarantees (CLAP~\citep{taeb2022provable}). These advances collectively enhance transparency, support counterfactual reasoning, and address concept leakage.

While existing methods effectively capture either causal relationships or correlations between factors, they rarely address both simultaneously. Our work focuses on achieving fine-grained controllable generation by accurately modeling both causal and correlational relationships within a unified framework, ensuring that generated data adheres to underlying generative mechanisms.

\section{Details of Experiments}

\subsection{Datasets, Baselines and Metrics}
\label{app:exp}

\paragraph{Datasets.}
\textbf{(1) Pendulum dataset} was originally proposed to explore causality~\citep{yang2021causalvae}. It contains about 7k images featuring 3 entities (\textit{pendulum}, \textit{light}, \textit{shadow}) and 4 concepts (\textit{Pendulum angle}, \textit{Light position}) → (\textit{Shadow position}, \textit{Shadow length}). To introduce correlation, we add a concept, \textit{Time}, which is correlated with \textit{Light position}.
\textbf{(2) Flow dataset} was proposed to explore causality~\citep{yang2021causalvae}. It contains about 8k images featuring 4 concepts (\textit{Ball size} → \textit{Water height}, (\textit{Water height}, \textit{Hole}) → \textit{Water flow}). 
\textbf{(3) dSprites dataset} was originally proposed to explore disentanglement~\citep{dsprites17}. It contains about 730k images featuring 6 independent concepts, from which we select \textit{scale}, \textit{x position}, and \textit{y position} to construct 4 concepts (\textit{scale}, \textit{x position}, \textit{x+y position}, \textit{$x^2$+$y^2$ position}) to explore correlation and possible causality. 
\textbf{(4) CelebA dataset} is a widely used real-world dataset in a range of computer vision tasks~\cite{liu2015deep}. It consists of 200k human face images with labels on various concepts. For our study, we focus on five concepts: \textit{Young}, \textit{Gender}, \textit{Receding hairline}, \textit{Makeup}, and \textit{Eye bag}, which are used to study causality and correlation.

\paragraph{Baselines.}
\textbf{(1) Semi-VAE} pairs the latent variables with target concepts by minimizing the MSE between latent variables and target concepts~\citep{kingma2014semi}; 
\textbf{(2) CSVAE} correlates a subset of latent variables with concepts by minimizing the mutual information~\citep{klys2018learning}; 
\textbf{(3) PCVAE} implements an invertible mapping between each pair of latent variables and concepts by enforcing the mutual dependence~\citep{guo2020property}. 
\textbf{(4) CorrVAE} uses a mask pooling layer to maintain the correlated mutual dependence between latent variables and concepts~\citep{wang2022multi}. 
\textbf{(5) ICM-VAE} learns disentangled causal representations by modeling independent causal mechanisms using nonlinear flow-based functions and a causal disentanglement prior supervised by causally related labels~\cite{komanduri2023learning}.
\textbf{(6) CausalVAE} learns causal representations of disentangled factors of data concepts via an SCM~\citep{yang2021causalvae}.
Additionally, we consider another method for the ablation study. 
\textbf{(7) C$^2$VAE-true} uses the ground-truth causal graphs and correlation masks. 

\paragraph{Metrics.}
For a comprehensive quantitative evaluation, we apply multiple metrics to measure data generation quality, concept reconstruction accuracy, and the mutual dependence between latent variables and concepts of interest. 
\textbf{(1) Data generation quality:} We use Frechet Inception Distance (\textbf{FID})~\citep{heusel2017gans} and Peak-Signal-to-Noise Ratio (\textbf{PSNR}) to assess the quality of generated images. FID compares the distribution of generated images with the distribution of a set of ground truth images. PSNR measures the difference between the original image and a reconstructed version. 
\textbf{(2) concept reconstruction accuracy:} We compute the Mean Absolute Error (\textbf{MAE}) between the model-predicted concept values and their ground truth labels. 
\textbf{(3) Mutual dependence:} We use the average mutual information (\textbf{avgMI})~\citep{locatello2019disentangling}, calculated as the Frobenius norm of the mutual information matrix of latent variables and concepts and the ground-truth mask matrix. 

\paragraph{Implementation Details.}
We implement C$^2$VAE using PyTorch, with key modules including a convolutional encoder and decoder (\texttt{ConvEncoder}, \texttt{ConvDec}), a differentiable DAG layer (\texttt{DagLayer}) for causal structure learning, and attention and masking modules to refine latent representations. The latent space is factorized into $n_{latent}$ latent variables, where $n_{latent}=n+k$ ($n$ is the number of targeted concepts and $k$ is the number of non-targeted concepts), each of dimension $d$, yielding a total latent dimension $l_{\text{dim}} = n_{latent} \times d$. In our experiments, we set $n_{latent} = 8$, $d = 4$. For each targeted concept, a mask pooling layer aggregates relevant latent variables, followed by a multi-layer perceptron (MLP) to predict the property value. The model is trained on 4 $\times$ NVIDIA RTX A4000 GPUs using AdamW~\citep{loshchilov2017decoupled} with learning rate $1\text{e-}3$ and weight decay $1\text{e-}3$.

% If so, you can uncomment the following line and ajust the path to include it.fig

\end{document}